\documentclass{article}
\usepackage{float}

\usepackage[final,sglblindworkshop]{neurips_2026}
\usepackage{graphicx}
\usepackage[T1]{fontenc}
\usepackage{hyperref}
\usepackage{url}
\usepackage{booktabs}
\usepackage{amsfonts}
\usepackage{nicefrac}
\usepackage{microtype}
\usepackage{xcolor}
\usepackage{subcaption}
\usepackage{placeins} 

\newcommand{\dn}{\,{\scriptsize$\downarrow$}}
\newcommand{\up}{\,{\scriptsize$\uparrow$}}

\workshoptitle{BiAlign: Dynamic Alignment in Human-AI Coupled Systems}
\title{Safety Side Effects of Preventative Persona Steering: An Empirical Audit}
\author{%
    Sabrina Tang \\
    University of California, Irvine \\
    \texttt{sabrit4@uci.edu} \\
    \And
    Nikolai Vo \\
    University of Toledo \\
    \texttt{hvo3@rockets.utoledo.edu} \\
    \And
    Carson Murillo \\
    University of California, San Diego \\
    \texttt{cemurillo@ucsd.edu} \\
    \And
    Chira Sukumarnpiti \\
    Berkeley City College \\
    \texttt{30056280@cc.peralta.edu} \\
    \And
    Sarah Wilson \\
    Google \\
    \texttt{sarahawilson@google.com}
}
\begin{document}
\maketitle
\begin{abstract}
High-level behaviors such as sycophancy and hallucination correspond to directions in a model's activation space \citep{chen2025persona}. These directions, known as \textit{persona vectors}, can suppress undesirable traits either at inference time, subtracting the vector during generation, or through preventative steering, adding the vector during fine-tuning, so the optimizer never needs to encode the trait in weights, then removing the vector at evaluation. \citet{chenarditi2025followup} recommend preventative steering because it preserves Massive Multitask Language Understanding (MMLU) accuracy, but validations of this recommendation have measured capability alone. We primarily evaluate results on preventative steering of sycophancy and hallucination vectors on three models, measuring three safety benchmarks alongside trait expression, coherence, and MMLU. Our findings show that preventative steering still changes refusal behavior, but by an amount that depends on the trait and does not establish that preventative persona steering is safety-neutral, and steering methods should be evaluated on several safety benchmarks.
\end{abstract}
\section{Introduction}
\label{sec:intro}
Persona vectors are directions in activation space for traits such as sycophancy and hallucination \citep{chen2025persona}, and recent work has shifted from subtracting them at inference to \emph{preventative steering} during fine-tuning. \citet{chenarditi2025followup} favor preventative steering for its capability retention but do not evaluate refusal safety, although activation steering is known to have unintended safety effects \citep{steeringexternalities,steeringsafety,safetypitfalls}. We audit preventative steering of the two traits on \texttt{Qwen2.5-7B-Instruct} and \texttt{Llama-3.1-8B-Instruct}, the two models of \citet{chenarditi2025followup}, and on \texttt{Qwen2.5-14B-Instruct} to test scale. We ask whether it changes refusal robustness at coefficients where MMLU, coherence, and trait suppression all look successful (\textbf{RQ1}), whether safety benchmarks agree on the direction of change (\textbf{RQ2}), and whether the effect is consistent across traits (\textbf{RQ3}) and across model families and scales (\textbf{RQ4}).

We evaluate three conditions (Section 3) on all three models, focusing on preventative steering.
\paragraph{Steering as a source of safety side effects.} Steering directions chosen to modify one behavior can alter refusal or weaken safety guardrails even when the intervention succeeds on its intended target \citep{steeringexternalities, steeringsafety, safetypitfalls}. These studies examine inference-time steering without persona vectors; to our knowledge, the interaction between preventative persona steering and refusal robustness has not been studied directly.

\paragraph{Layer choice and reliability.}
\citet{heyman2026steerlikellmactivation} find that steering effects vary substantially with layer and token position, and \citet{steeringreliability} identify sensitivity to steering coefficients and prompts as key reliability concerns. We adopt their layer choices and follow their advice to sweep coefficients, but extend this work in two ways: we study steering applied during fine-tuning rather than at inference, and we judge reliability by its side effects on refusal across three safety benchmarks rather than by success on the targeted behavior.
\section{Methodology}
\label{sec:method}
\paragraph{Persona vectors.} We extract sycophancy and hallucination vectors with the unmodified pipeline of \citet{chen2025persona}, so safety differences reflect how the vector is applied, not how it was derived.

\paragraph{Conditions.} For each trait and model we compare three conditions, sweeping the steering coefficient $c \in \{0, 0.25, \ldots, 3\}$ for the two steered conditions. The \textbf{fine-tuned base model} is trained on the trait data with no steering, isolating the effect of steering from that of fine-tuning. \textbf{Preventative steering} adds the vector to the residual stream during the same fine-tuning via a forward hook. Note that preventative steering at $c = 0$ is equivalent to the fine-tuned base model. \textbf{Inference-time steering} subtracts the vector from the residual stream of the original Instruct model at generation, with no fine-tuning or weight change.

Following \citet{chen2025persona}, we apply inference-time steering to the original Instruct model, since the steering vectors are extracted from its activations and this isolates the vector's causal effect from fine-tuning-induced changes. We steer one residual-stream layer per model and trait, following \citet{chen2025persona} and \citet{heyman2026steerlikellmactivation}: layer 20 (sycophancy) and layer 16 (hallucination) on \texttt{Qwen2.5-7B-Instruct}, and layer 16 for both traits on \texttt{Llama-3.1-8B-Instruct}. Layer indices start from 1, following \citet[App.~C]{heyman2026steerlikellmactivation}. Neither source covers \texttt{Qwen2.5-14B-Instruct}, so we scale the \texttt{Qwen2.5-7B-Instruct} layers by relative depth (28 to 48 layers), steering at layer 34 for sycophancy and layer 27 for hallucination. Both steering methods use the same layer.

\paragraph{Evaluation.} For preventative steering we measure, at every coefficient, attack success (HarmBench ASR \citep{harmbench2024}, StrongREJECT score \citep{strongreject2024}), compliance on unsafe prompts from XSTest \citep{xstest2023}, trait expression and coherence (persona-vector scoring pipeline \citep{chen2025persona}), and MMLU; for inference-time steering we measure trait expression and coherence. HarmBench responses are labeled by the official HarmBench classifier (HarmBench-Llama-2-13b-cls), StrongREJECT responses are scored with its rubric by GPT-4o-mini, and XSTest responses are labeled by GPT-4 as full compliance, partial refusal, or full refusal. We spot-check judge labels by hand to separate harmful compliance from noisy or incoherent output.

\section{Results}
\label{sec:results}
\begin{figure}[!t]
  \centering
  \includegraphics[width=0.9\linewidth]{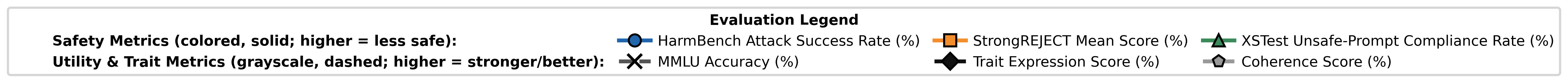}
  \includegraphics[width=0.8\linewidth]{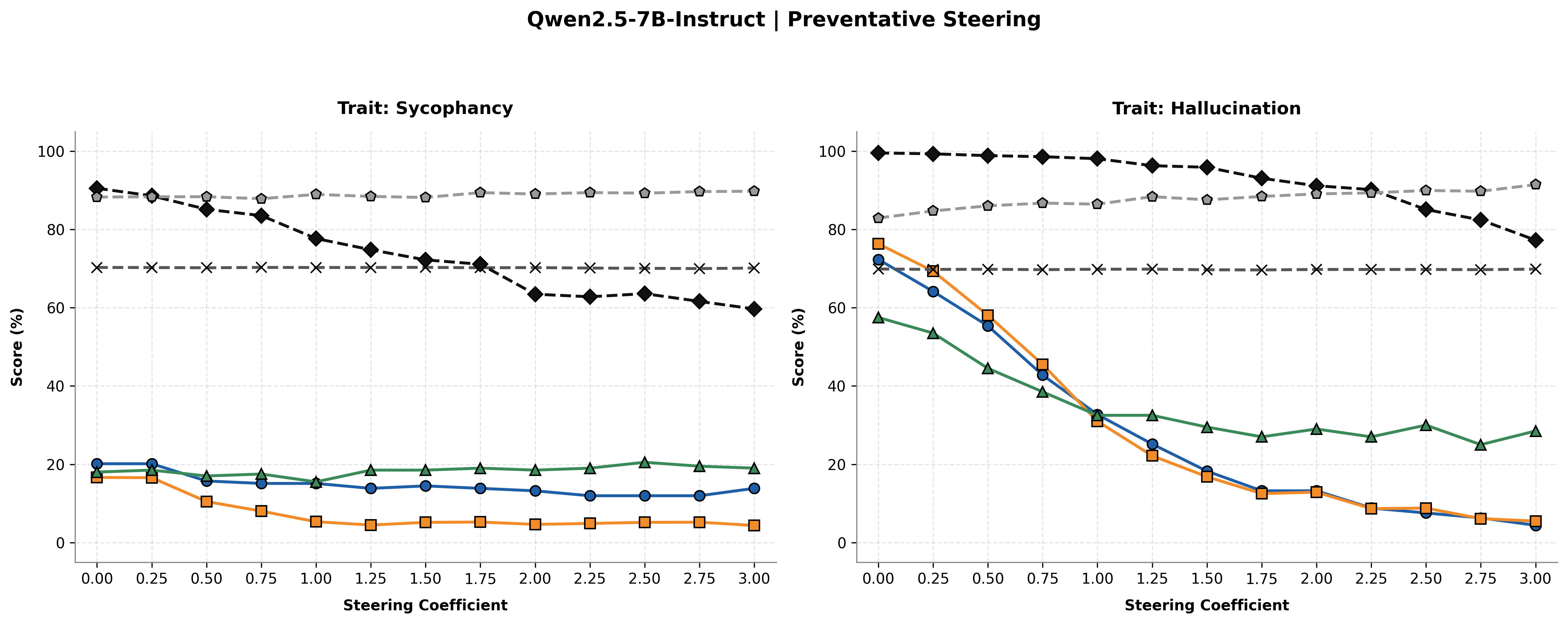}\\[2pt]
  \includegraphics[width=0.8\linewidth]{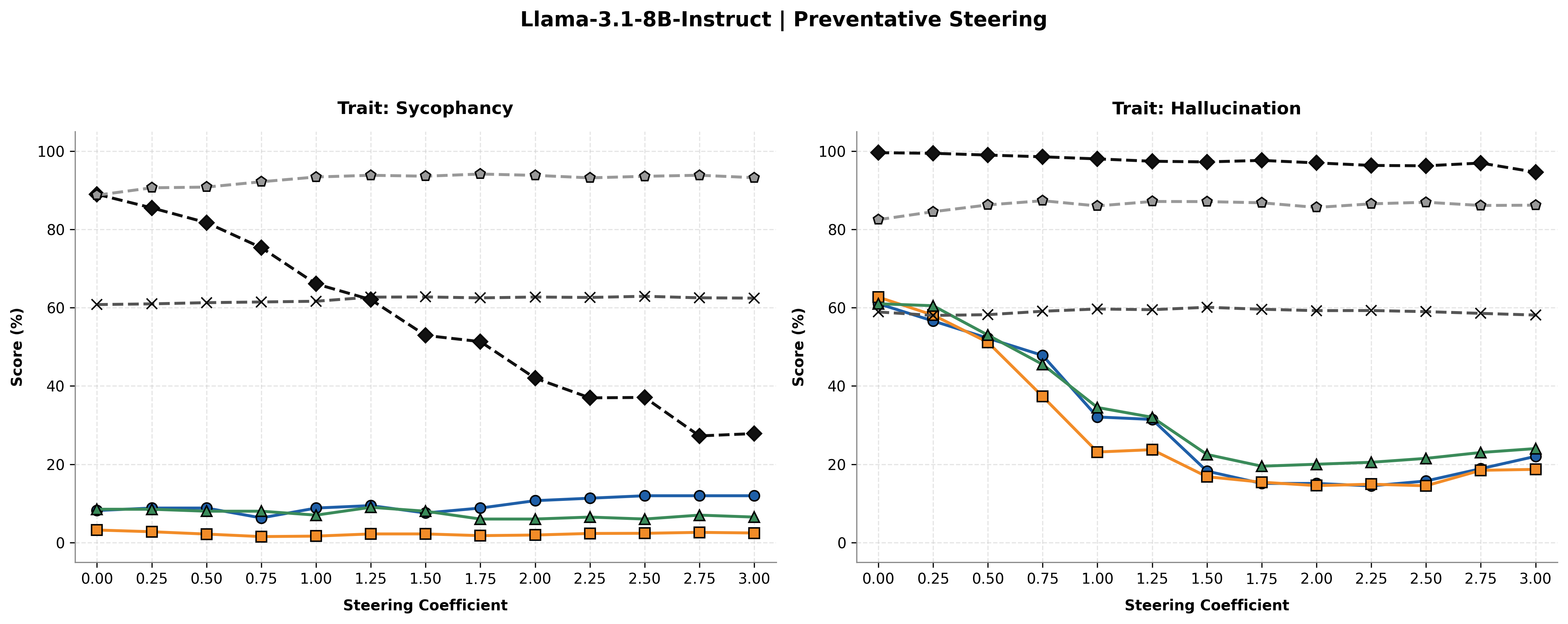}\\[2pt]
  \includegraphics[width=0.8\linewidth]{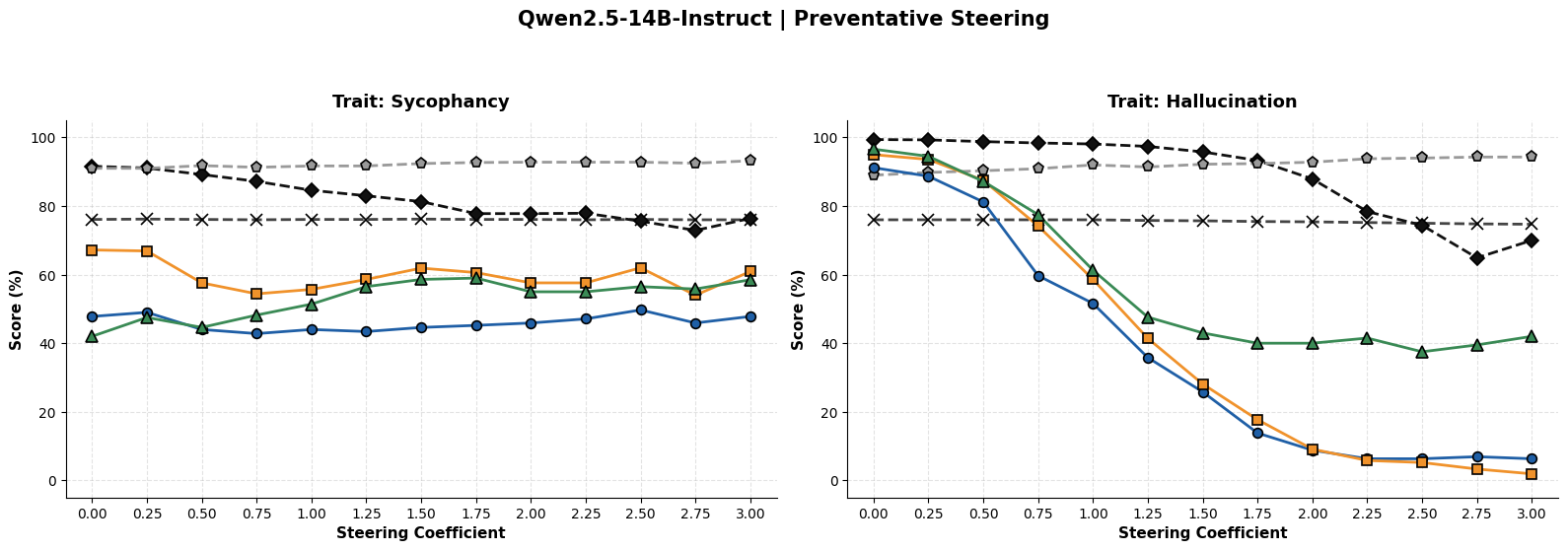}\\[2pt]
  \caption{Preventative steering coefficient sweeps for \texttt{Qwen2.5-7B-Instruct} (top), \texttt{Llama-3.1-8B-Instruct} (middle), and \texttt{Qwen2.5-14B-Instruct} (bottom), for sycophancy (left) and hallucination (right). Colored solid lines are safety metrics (higher = less safe); grayscale dashed lines are utility and trait metrics (higher = more accurate, stronger trait, or more coherent).}
  \label{fig:sweeps}
\end{figure}

\paragraph{Hallucination steering shifts safety far more than the target trait (RQ1--RQ3).}On all three models, preventative hallucination steering sharply lowers attack success. From $c=0$ to $c=3$, HarmBench ASR falls from 61.0\% to 22.0\% on \texttt{Llama-3.1-8B-Instruct}, from 72.3\% to 4.4\% on \texttt{Qwen2.5-7B-Instruct}, and from 91.2\% to 6.3\% on \texttt{Qwen2.5-14B-Instruct}, and StrongREJECT follows the same downward trend. Trait expression drops much less over the same range, staying at 70\% or above on every model (Table~\ref{tab:conditions}). Most of the safety shift happens at $c \le 1.25$, where trait expression is still above 96\% on every model. The hallucination vector thus acts on refusal behavior \emph{more strongly than on hallucination itself}, a coupling that capability and trait metrics would not reveal.

Two results qualify this pattern. First, the effect is non-monotonic on Llama: ASR reaches its minimum of 14.5\% at $c=2.25$ and rises again to 22.0\% by $c=3$. Second, XSTest does not follow the other two benchmarks. While HarmBench and StrongREJECT approach zero on both Qwen models, XSTest unsafe compliance levels off at roughly 25--30\% on \texttt{Qwen2.5-7B-Instruct} and 37--43\% on \texttt{Qwen2.5-14B-Instruct}, so steering restores refusal on the attack types those benchmarks cover but not on all unsafe requests.

Figure~\ref{fig:sweeps} shows the preventative-steering sweeps. Coherence is flat or slightly rising across every sweep, and MMLU varies by at most 2.1 percentage points within any sweep: 69.6--70.3\% on \texttt{Qwen2.5-7B-Instruct}, 58.0--62.9\% on \texttt{Llama-3.1-8B-Instruct}, and 74.7--76.2\% on \texttt{Qwen2.5-14B-Instruct}. None of the safety shifts below therefore coincide with capability degrading.

\paragraph{Sycophancy steering produces small shifts on which benchmarks disagree (RQ2, RQ4).}
Sycophancy steering suppresses the trait more (Table~\ref{tab:conditions}) but moves safety metrics only modestly, and the direction depends on both model and benchmark. On \texttt{Llama-3.1-8B-Instruct}, HarmBench ASR rises from 8.2\% to 11.9\% (a 46\% relative increase) while XSTest unsafe compliance falls from 8.5\% to about 6.5\%. On \texttt{Qwen2.5-7B-Instruct}, HarmBench and StrongREJECT both fall (20\% to 14\%, 17\% to 5\%, respectively) while XSTest compliance rises slightly. On \texttt{Qwen2.5-14B-Instruct}, HarmBench ends where it starts (48\%), StrongREJECT falls from 67\% to 61\%, and XSTest compliance rises from 42\% to 59\%. In every model, at least one benchmark moves opposite to another, so no single benchmark would have characterized the effect.

\paragraph{Fine-tuning data sets the safety baseline.}
At $c=0$, attack success differs sharply by trait on the same base model: HarmBench ASR is 8\% vs.\ 61\% on Llama, 20\% vs.\ 72\% on Qwen-7B, and 48\% vs.\ 91\% on Qwen-14B for sycophancy vs.\ hallucination data. Fine-tuning on the hallucination data appears to erode refusal substantially on its own, and much of what preventative steering does on that trait may be to counteract this erosion.

\begin{table}[H]
  \caption{Trait expression (Tr.) and coherence (Coh.), in \%, at $c=0$ and $c=3$. Arrows show the direction of change from $c=0$ to $c=3$. Inference-time steering at $c=0$ is the original Instruct model; preventative steering at $c=0$ is the fine-tuned base model.}
  \label{tab:conditions}
  \centering
  \small
  \setlength{\tabcolsep}{4pt}
  \begin{tabular}{llrrrrrrrr}
    \toprule
    & & \multicolumn{4}{c}{Inference-time steering} & \multicolumn{4}{c}{Preventative steering} \\
    \cmidrule(lr){3-6}\cmidrule(lr){7-10}
    & & \multicolumn{2}{c}{$c=0$} & \multicolumn{2}{c}{$c=3$} & \multicolumn{2}{c}{$c=0$} & \multicolumn{2}{c}{$c=3$} \\
    \cmidrule(lr){3-4}\cmidrule(lr){5-6}\cmidrule(lr){7-8}\cmidrule(lr){9-10}
    Model & Trait & Tr. & Coh. & Tr. & Coh. & Tr. & Coh. & Tr. & Coh. \\
    \midrule
    Qwen2.5-7B   & Syc. &  5.1 & 99.0 & 0.2\dn & 93.7\dn & 90.5 & 88.3 & 59.8\dn & 89.7\up \\
                 & Hal. & 17.8 & 92.7 & 0.9\dn & 95.3\up & 99.5 & 83.0 & 77.3\dn & 91.5\up \\
    Llama-3.1-8B & Syc. &  4.0 & 97.8 & 1.0\dn & 89.4\dn & 88.9 & 89.2 & 27.9\dn & 93.2\up \\
                 & Hal. & 34.5 & 73.4 & 5.3\dn & 39.7\dn & 99.6 & 82.5 & 94.6\dn & 86.2\up \\
    Qwen2.5-14B  & Syc. &  3.9 & 99.4 & 0.2\dn & 95.3\dn & 91.6 & 91.0 & 76.4\dn & 93.2\up \\
                 & Hal. &  6.4 & 96.5 & 0.7\dn & 94.9\dn & 99.4 & 89.0 & 70.0\dn & 94.3\up \\
    \bottomrule
  \end{tabular}
\end{table}

\section{Discussion and Limitations}
\label{sec:discussion}
Persona vectors may be partly entangled with refusal behavior, and evaluating them should include several safety benchmarks alongside capability and target-trait metrics. In our runs this entanglement mostly improved safety, but the Llama sycophancy result shows it need not.

Table~\ref{tab:conditions} compares the conditions. Fine-tuning on the trait data raises trait expression from 4--34\% in the original Instruct models to 89--100\%. Because inference-time steering was applied to the original models, its sweep never reaches the trait range of the fine-tuned models, so the two methods cannot be calibrated to equal trait expression. Inference-time steering pushes trait expression below 6\% on every model with coherence above 89\%, except for Llama hallucination, where coherence falls to 40\% at $c=3$; preventative steering keeps coherence flat but only partially suppresses the fine-tuned trait.

Each point is a single run, the sweep is coarse, and confidence intervals are not shown, which limits claims about dose-response shape. The \texttt{Qwen2.5-14B-Instruct} layers are scaled from the 7B choices rather than selected by a sweep, so its results may partly reflect a suboptimal layer. HarmBench, StrongREJECT, and XSTest probe different aspects of safety, and our findings may not extend to other benchmarks, adaptive attacks, or model families.

\section{Conclusion}
\label{sec:conclusion}
We compared three conditions (the fine-tuned base model, preventative steering, and inference-time steering) across two traits and three models. Although MMLU and coherence stay essentially unchanged, preventative steering still changes refusal behavior, by an amount that depends on the trait: hallucination steering lowers HarmBench ASR by 39--85 points on all three models, far more than it reduces the trait, while sycophancy steering produces small shifts whose direction depends on the model and on which safety benchmark measures it. Capability retention therefore does not establish that preventative persona steering is safety-neutral, and steering methods should be evaluated on several safety benchmarks.

\bibliographystyle{plainnat}
\bibliography{references}
\end{document}